\documentclass[runningheads]{llncs}
\usepackage{graphicx}

\usepackage{subfigure}

\usepackage{amsmath}
\usepackage{amssymb}
\usepackage{booktabs}
\usepackage{caption}
\usepackage{subcaption}
\usepackage{bm}
\usepackage{amssymb}
\usepackage{amsmath}
\usepackage{multirow}

\begin{document}
\title{MovePose: A High-performance Human Pose Estimation Algorithm on Mobile and Edge Devices}
%
%

\author{Dongyang Yu-{yudongyang2022@gmail.com}
Haoyue Zhang-{hz625@cornell.edu}
Ruisheng Zhao-{rathenzrs@gmail.com}
Guoqi Chen-{276851182@qq.com}
Wangpeng An-{anwangpeng@gmail.com}
Yanhong Yang-{yyh@cueb.edu.cn}}

\institute{}
%

%

%
\maketitle              
\begin{abstract}
We present MovePose, an optimized lightweight convolutional neural network designed specifically for real-time body pose estimation on CPU-based mobile devices. The current solutions do not provide satisfactory accuracy and speed for human posture estimation, and MovePose addresses this gap. It aims to maintain real-time performance while improving the accuracy of human posture estimation for mobile devices. Our MovePose algorithm has attained an Mean Average Precision (mAP) score of 68.0 on the COCO \cite{cocodata} validation dataset. The MovePose algorithm displayed efficiency with a performance of 69+ frames per second (fps) when run on an Intel i9-10920x CPU. Additionally, it showcased an increased performance of 452+ fps on an NVIDIA RTX3090 GPU. On an Android phone equipped with a Snapdragon 8 + 4G processor, the fps reached above 11. To enhance accuracy, we incorporated three techniques: deconvolution, large kernel convolution, and coordinate classification methods. Compared to basic upsampling, deconvolution is trainable, improves model capacity, and enhances the receptive field. Large kernel convolution strengthens these properties at a decreased computational cost. In summary, MovePose provides high accuracy and real-time performance, marking it a potential tool for a variety of applications, including those focused on mobile-side human posture estimation. The code and models for this algorithm will be made publicly accessible.

\keywords{Human Pose Estimation  \and Accuracy \and Edge devices.}
\end{abstract}
%
%
%

\section{INTRODUCTION}

The era of smart devices and edge computing has ushered in a time of rapid digitization, transforming the face of technology and computation. These evolutionary strides are increasingly shaping our lives with a growing number of applications from self-driving cars, smart home automation to healthcare monitoring systems. One such application that has garnered significant attention of late is human pose estimation - the process of inferring the pose or the orientation of the human body using computational algorithms deployed on imaging devices. Brought to the fore with the rising interest in augmented reality (AR), robot-human interaction, activity recognition, and gaming, having an efficient and accurate human pose estimation system is becoming paramount.

However, performing human pose estimation on edge devices, CPUs, and mobile devices introduces unique challenges. These platforms face resource constraints, including limited computing power, memory, and energy, which can severely impede the execution of complex pose estimation algorithms needed for real-time applications. Furthermore, issues relating to privacy and latency from cloud-based solutions necessitate the need for more localized, device-centric solutions.

\begin{figure}
  \centering
  \includegraphics[height=1.8in,width=3.6in]{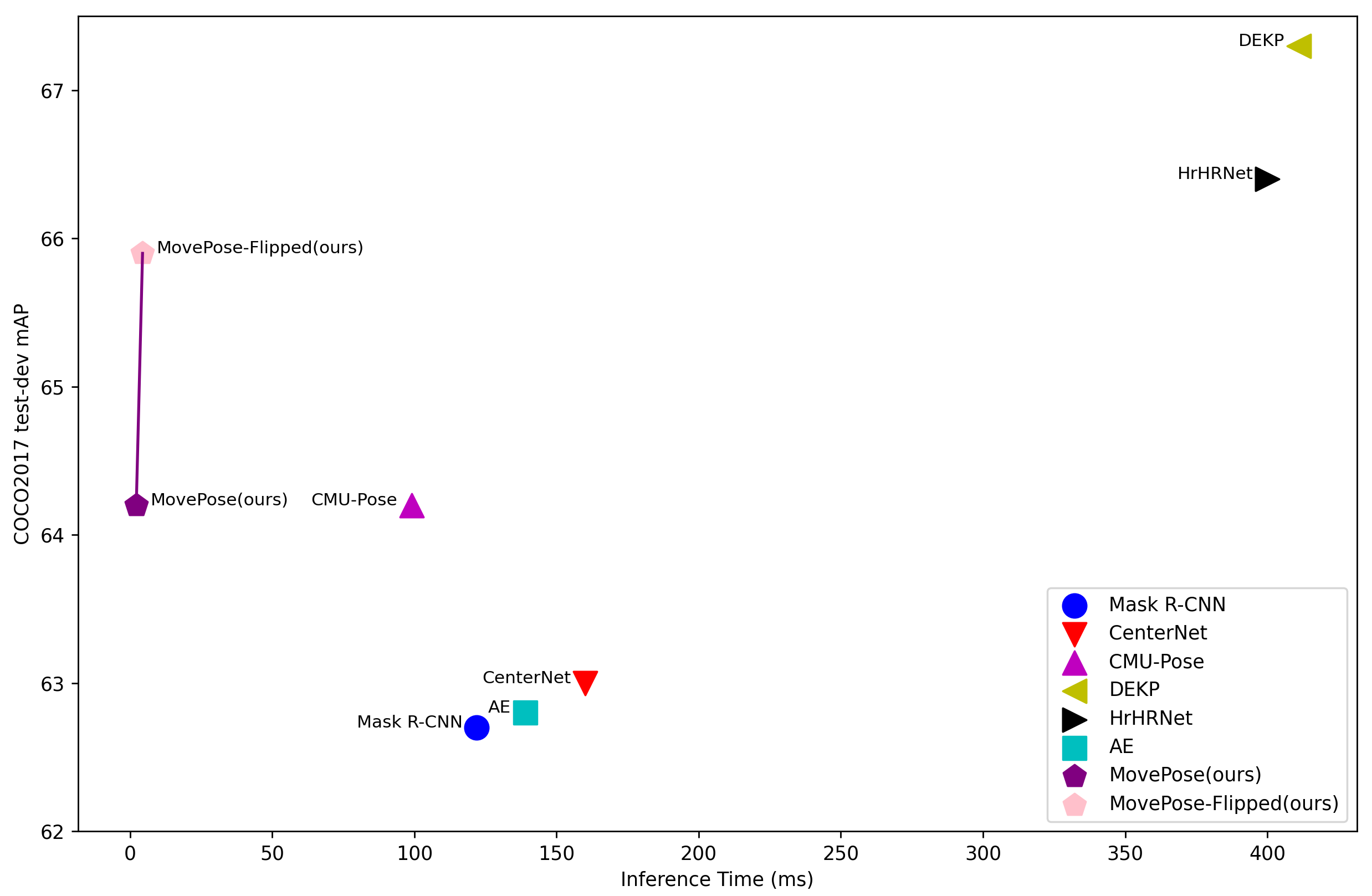}  

  \caption{A comparison of the MovePose method with other established methods, based on mAP and inference time, using the COCO2017 test-dev dataset.}
  \label{fig:overview1}
\end{figure}

Numerous algorithms, each with its unique attributes and shortcomings, have been proposed within the realm of human pose estimation. VitPose~\cite{vitpose}, a cutting-edge model representing the top-down algorithm category, employs a transformer network structure. Regrettably, this structure does not lend itself effectively to accelerated inference on edge devices. JCRA~\cite{jcra} , a champion of the bottom-up domain, provides high keypoint recognition accuracy and a marked increase in speed over OpenPose~\cite{realtime}. Yet, similar to VitPose~\cite{vitpose}, JCRA's transformer network structure is not optimal for speedy inference on edge or mobile devices. While BlazePose~\cite{blazepose}, a top-down human pose estimation algorithm, attains real-time human pose-estimation in terms of operational speed, its keypoint recognition precision leaves much to be desired. Real-time high-precision multi-person human pose estimation on edge devices is an urgent problem to be solved.

In response to these challenges, we present MovePose, a high-performance algorithm for human pose estimation designed explicitly for deployment on edge, CPU, and mobile devices. Movepose is designed leveraging large convolutions to expand the receptive field, scrutinizing a broader area of feature maps, and thereby acquiring superior global features. This algorithm also uses a deconvolution network as a substitute for the usual bicubic interpolation upsampling approach, simplifying the complexity of estimation along with an increase in precision.

Furthermore, MovePose exploits the SimCC \cite{simcc} strategy, reconceptualizing pose estimation from a traditional regression task into a classification task. This conversion significantly reduces computational demands and accelerates processing speed. The transformation of the estimation task into a binary classification problem within the predefined proximity of the target joint decreases its complexity and enhances the accuracy. Movepose demonstrates the advantage of balance between inference time and accuracy in Fig. \ref{fig:overview1}.

The main contributions of this paper are:
\begin{itemize}
    \item Addressing a current gap in the performance and accuracy of real-time body pose estimation on mobile devices. We introduced MovePose, an optimized lightweight convolutional neural network specifically designed for this task. This network not only maintains real-time performance but also improves the accuracy of human posture estimation on mobile devices, a result not seen in existing solutions.
    \item Enhancing the accuracy of keypoint detection through the incorporation of three techniques: deconvolution, large kernel convolution, and coordinate classification methods. Compared to basic upsampling, deconvolution is trainable and improves the model's capacity and receptive field. Large kernel convolution further strengthens these properties at a decreased computational cost. 
    \item  Achieving remarkable performance on the COCO validation dataset with a 68.0 mAP. MovePose is suitable for real-time applications such as fitness tracking, sign language interpretation, and advanced mobile human posture estimation, highlighting its potential as an influential tool in the field. Furthermore, we are committed to enhancing transparency and reproducibility by making our code and model available to the public.

\end{itemize}

\section{Related Works}
Human pose estimation(HPE) has seen an unprecedented surge in research interest, owing to its vast application spectrum spanning augmented reality, video animation, health monitoring systems, and beyond. This section will trace the trajectory of advancements in this field, paying particular attention to the realms of heatmap-based techniques, top-down methods, bottom-up methods, and the SimCC 
 \cite{simcc}approach.
\subsection{Top-Down Methods}
Top-down pose estimation algorithms operate in two distinct phases. The initial phase involves the detection of human bounding boxes, which is followed by the estimation of a lone individual's pose within each identified bounding box. Prototypical models incorporating this approach, including the likes of AlphaPose \cite{rmpe2017}, CPN \cite{cpn2017}, SimpleBaseline \cite{BinXiao2018SimpleBF}, HRNet \cite{wang2020deep}, and Vitpose \cite{vitpose}, standardize the process by instituting the detection of single individual keypoints inside a bounding box.

These algorithms conventionally generate person bounding boxes primarily via an object detector. Furthering this technique, Mask R-CNN \cite{mask} incorporated a keypoint detection branch within the Faster R-CNN \cite{faster}, resulting in the efficient reuse of features post-ROIAlign. An attribute ubiquitous to all mentioned algorithms is the utilization of heatmaps for the encoding and decoding of keypoints.

Despite the high accuracy of top-down methods, their heavy dependence on detector performance is noteworthy. While the incorporation of a personal detector enhances performance, greater run-time is incurred, culminating in a practice that is cost-inefficient.

\subsection{Bottom-Up Methods}

The field of HPE has seen significant advancements with the introduction of bottom-up algorithms. These algorithms, such as CMU-Pose \cite{realtime}, identify and associate individual body parts to their respective individuals in images. CMU-Pose \cite{realtime}, a real-time pose estimation model, has pioneered this approach by identifying all body joints independently and integrating them into individual poses. However, its high computational and energy demands pose challenges for deployment on resource-limited devices.
Associative Embedding, another bottom-up method, uses a Convolutional Neural Network (CNN) to predict body joints and their associated "tag maps," connecting the joints based on spatial and similarity analysis of these maps. AlphaPose \cite{rmpe2017}, a real-time multiperson pose estimation system, uses a Region Proposal Network (RPN) for individual detection and a separate network for pose estimation. The Stacked Hourglass Network \cite{stacked}, another significant contribution, effectively determines spatial relationships between body parts by consolidating features of varying scales.
Despite their computational efficiency, bottom-up algorithms can struggle with accuracy in complex scenarios, such as crowded environments or uncommon poses, and may incorrectly associate body parts with individuals. Their high computational resource requirements also limit their applicability for real-time applications or on resource-constrained devices. These challenges highlight the need for ongoing research to improve and optimize these computation-intensive algorithms.

\subsection{2D Heatmap-Based Techniques}
In the realm of 2D human pose estimation, numerous strategies, especially heatmap-based approaches, have significantly contributed to the field's advancement. This article critically examines these techniques, assessing their design principles, methodologies, and performance capabilities.

DeepPose \cite{toshev2014deeppose}, a pioneering approach, signified the onset of deep learning applications in human pose estimation. It employs a series of pose regressors to predict body joint positions and further refines them using image patch-based regression.

Building on this foundation, Convolutional Pose Machines (CPMs) \cite{ShihEnWei2016ConvolutionalPM} employed a multi-stage architectural set-up to facilitate intermediate supervision. This architectural plan enabled the learning of complex dependencies, substantially improving human pose prediction.

OpenPose \cite{realtime}introduced a real-time technique using Part Affinity Fields (PAFs) for pose estimation. This technique centered on real-time pose estimation, setting a benchmark for real-time applications.

Despite these advancements, these methods' performance often suffers due to the limited and low resolution of heatmaps, signaling the need for enhanced solutions, which this article aims to explore.

\subsection{Coordinate Classification}
In the relevant works of Human Pose Estimation, the application of Coordinate Classification presents a fresh perspective, such as embodied in the approach named SimCC \cite{simcc}. Traditionally, the estimation of human gestures in a single image was perceived as a regression problem. In this standard practice, the task's difficulty arises from the diverse human postures and the occasional occlusion of body joints.

Contrarily, SimCC \cite{simcc} treats this task as a multi-label classification problem, disrupting the conventional methods. Each coordinate of the human joints on an image is mapped onto class labels, these labels are then learned using a simplistic yet effective classification scheme. This innovative approach manages to carve out substantial improvements in the performance of HPE tasks.

Moreover, SimCC's methodology inherently contains spatial context information, providing it with the capacity to manage complex poses and instances of multiple persons sans the requirement of explicit reasoning mechanisms. This advantage amplifies the model's utility and elucidates the methodological progress in the HPE studies, precipitating future works' potential interpretability and simplicity.

\begin{figure}
  \centering
  \includegraphics[height=1.6in, width=\linewidth]{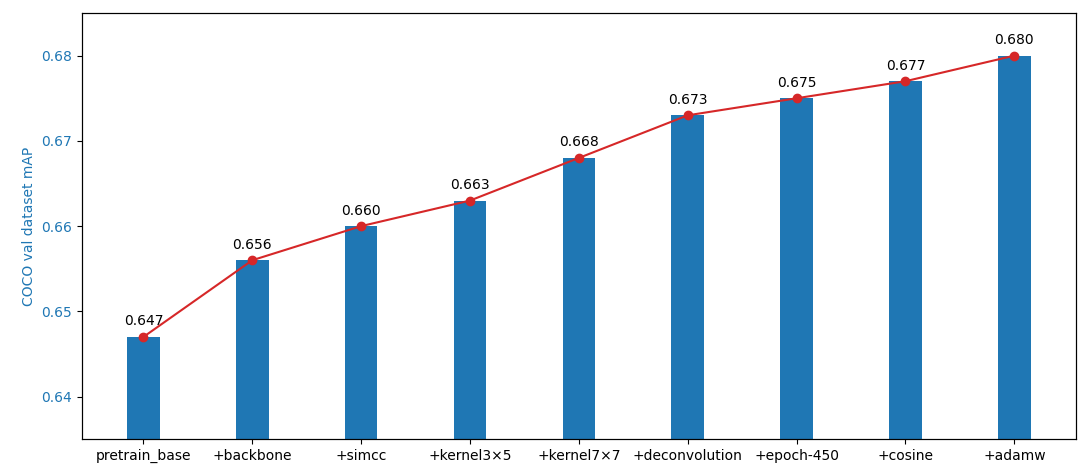}  

  \caption{Ablation study showing the improvement in Mean Average Precision (mAP) for MovePose on the COCO validation dataset for each deep learning enhancement technique.}
  \label{fig:Ablation}
\end{figure}

\section{Algorithm and Techniques}
In this section, we delve into the methodologies and techniques adopted to address the challenges identified earlier. Each subsection provides an in-depth look into the specific components of our proposed solutions, detailing their architecture, functionality, and significance.

The Fig.~\ref{fig:Ablation} presents an ablation study conducted to examine various deep learning techniques, aiming to optimize and enhance the performance of MovePose in body pose estimation across different devices. The efficacy of each technique is evaluated using the Mean Average Precision (mAP) score on the COCO validation dataset as an accuracy metric.

\begin{figure*}
\begin{subfigure}{}
  \centering
\includegraphics[height=1.6in,width=\linewidth]{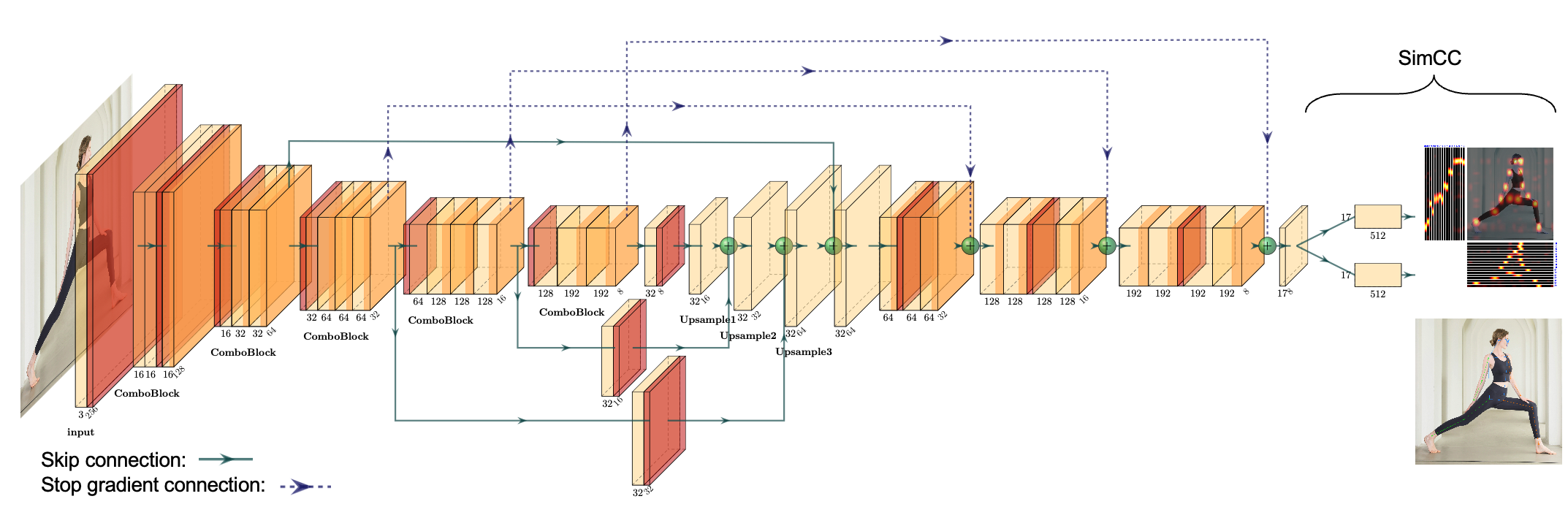}  

  \label{fig:movepose}
\end{subfigure}

\caption{Overview of MovePose algorithm}
\label{fig:overview2}
\end{figure*}

\subsection{Architecture of the Movepose Network}
Fig.~\ref{fig:overview2} illustrates the distinctive design of the MovePose network. It is a deep learning architecture specifically designed for human pose estimation tasks, fusing elements of U-Net~\cite{unet2015}, transposed convolution for upsampling, and the Simple Coordinate Classification (SimCC) \cite{simcc} technique to enhance the prediction accuracy. The MovePose uses MobileNet as its backbone. 

The MovePose uses both skip connections and upsampling strategies to integrate features from various levels and simultaneously capture high-level and low-level pose details across different dimensions. This approach moves beyond simple downsampling via the encoding layer followed by upsampling to restore the original resolution. Instead, the MovePose network depends on skip connections to maintain the high-resolution details that are essential for pose estimation.

The network's second key feature is the use of the transposed convolution, also known as deconvolution. This operation, termed ConvTranspose2d in PyTorch, allows the upsampling of lower-resolution feature maps in order to match the resolution of the image. This approach enhances the spatial accuracy of pose predictions.

Lastly, by utilizing the SimCC~\cite{simcc} technique, the MovePose network transforms the pose estimation problem into a spatial dimensions classification problem, which increases both the accuracy and consistency of the estimations.

\subsection{SimCC Method: An In-depth Exploration}

The SimCC method lays the groundwork for our approach. For a thorough understanding of the enhancements made by MovePose, we delve into SimCC with more depth.

SimCC~\cite{simcc}, standing for Simple Coordinate Classification, reconfigures the conventional representation of human joints. Typically, in the realm of human pose estimation, the location of a joint is deduced from the high-density regions of a heatmap. However, SimCC~\cite{simcc} proposes an alternate methodology, where it symbolizes human joints via a coordinate map instead of a heatmap.

This map locally classifies all plausible locations to exhibit predictions for body joints. In essence, SimCC~\cite{simcc} processes the given input image and produces a series of 2D maps. Each 2D map is linked to one body joint, and every pixel of this map is assigned a probability value, indicative of the likelihood of that pixel being the accurate position for the corresponding joint. 

For a specific keypoint type, we compute the final forecasted absolute joint position $(\hat{o}_x,\hat{o}_y)$ by leveraging the two $1$D vectors \bm{$o_x$} and \bm{$o_y$} that are generated by the model. The formula is as follows:
\begin{equation}
    \hat{o}_x = \frac{\mathop{\arg\max}_{i}{(\bm{o_x}(i))}}{k},\hat{o}_y = \frac{\mathop{\arg\max}_{j}{(\bm{o_y}(j))}}{k}.
\end{equation}

A splitting factor is referred to as $k$, which is always greater than or equal to one $k (\geq 1)$. 
Consequently, pose estimation transforms into a classification task in SimCC~\cite{simcc}, deviating from a regression task. This shift in perspective simplifies the intricacies involved in inference. The issue reduces to binary classification tasks within pre-established zones around the target joints. As an outcome, the model's precision is heightened, successfully circumventing challenges commonly associated with regression models.

\begin{table}[!htbp]
\scriptsize
\centering
\setlength{\tabcolsep}{10pt} 
\begin{tabular}{|c|c|}
\hline
\begin{tabular}[c]{@{}c@{}}Convolution\\ Kernel Size\end{tabular} & \begin{tabular}[c]{@{}c@{}}Mean Average Precision\\ (mAP) on COCO Validation Dataset\end{tabular} \\ \hline
3×3                                                               & 65.4                                                                                                   \\ \hline
5×5                                                               & 66.9                                                                                                   \\ \hline
5×5 + 7×7                                                         & 67.2                                                                                                   \\ \hline
7×7                                                               & 67.3                                                                                                   \\ \hline
\end{tabular}
\caption{Comparison of Mean Average Precision (mAP) results on the COCO validation dataset for models with different convolution kernel sizes.}
\label{tbm:kernel}
\end{table}

\subsection{Backbone Enhancements with Large Kernel Convolution}

Large kernel convolution is integrated into our network to capture detailed information and larger contextual spatial data within images. This approach primarily improves the model's predictive accuracy by allowing for a wider field of view and the ability to digest more complex patterns.
Convolutional layers utilizing 7x7 kernels exhibit larger receptive fields in contrast to those that employ 3x3 kernels. As a result, they can encapsulate a more inclusive information spectrum of the image.
Optimizing the kernel size is a crucial component to the successful implementation of convolutional neural networks (CNNs). The core idea behind employing a large kernel convolution is to capture a broader view of input images. Standard convolutional layers with smaller kernels may fail to encapsulate a wider context in the input. Conversely, those with a larger kernel are designed and optimized to gather more information from wider fields of view. This makes them particularly effective at managing spatial hierarchies in images - a significant challenge in image processing tasks.

The Table~\ref{tbm:kernel} displays four different kernel sizes: '3×5', '5×5', '5×5+7×7', and '7×7'. Relative to these kernel sizes, the mAP scores obtained were approximately 65.4, 66.9, 67.2, and 67.3, respectively. This shows a general increase in the mAP score as the kernel size was increased.

\subsection{Deconvolution Network: Mechanisms and Merits}

Mechanisms: The deconvolution procedure encompasses scaling-up of feature maps from a lower to a higher resolution. This task is accomplished via the padding of zeros to the input volume, followed by execution of a convolution operation. Consequently, the cumulation of these occurrences, specifically, the padded zeros and acquired weights yield an output with magnified dimensions.

\begin{table}[h]
\scriptsize
\centering
\setlength{\tabcolsep}{10pt} 
\begin{tabular}{|c|ccc|cc|}
\hline
Method&$AP^{kp}$&$AP_{50}^{kp}$&$AP_{75}^{kp}$&$AP_{M}^{kp}$&$AP_{L}^{kp}$\\
\hline
Upsample&64.0 &88.3& 71.6& 61.2 &68.2\\
\hline

Deconvolution &\textbf{64.7} &\textbf{88.4}& \textbf{71.8}& \textbf{61.8} &\textbf{69.1}\\
\hline

\end{tabular}%
\caption{Comparisons of the UpSample and Deconvolution methods on the coco val dataset.}
\label{tbm:deconvolution}
\end{table}

As opposed to Upsampling techniques like Pytorch's nn.Upsample, the weight in use during the deconvolution process serves as a learnable parameter during the network's training phase. Instead of directly expanding the dimensions of an image or volume via interpolation, as executed in Upsampling, deconvolution advances further by determining the optimal method for this amplification via back-propagation and gradient descent.

Merits: A principal merit of deconvolution networks entails empowering the model to efficaciously learn representation from data, consequently escalating the model's capacity. In deploying deconvolution, the model achieves superior detailing in its output by incorporating the subtle patterns discerned from the original input. Furthermore, the deconvolution networks possess the capability to extend the network’s receptive field, thereby capturing expansive contextual information.

As indicated in the Table~\ref{tbm:deconvolution}, Deconvolution method has performed better than the Upsample method on all tested parameters.

\begin{figure*}
  \centering
\includegraphics[height=1.6in,width=\linewidth]{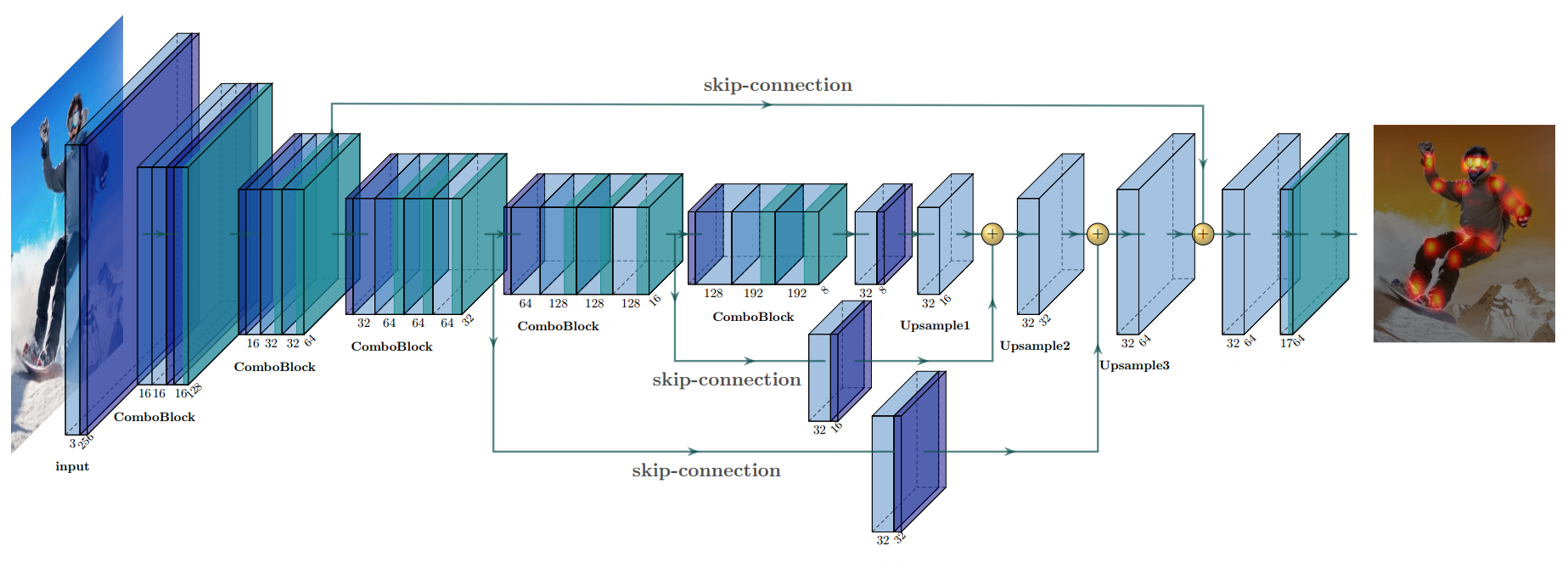}

\caption{Overview of Lite Model algorithm}
\label{fig:overview3}
\end{figure*}

\subsection{Lite Model Pre-training Approach}
The Lite model, as depicted in Fig. \ref{fig:overview3}, designed for pose estimation, uses a highly sophisticated yet efficient architecture to improve both accuracy and computational efficiency. It combines a heatmap regression approach – a standard in pose estimation methods – with a post-processing technique borrowed from the Distribution-Aware Coordinate Representation of Keypoint (DARK) \cite{dark_model} method. The Lite model has successfully obtained a mean average precision (mAP) of 64.7 on the COCO validation dataset.

The initial process of the model involves generating a heatmap, where the DARK \cite{dark_model}method implemented subsequently plays a crucial role. The objective of DARK \cite{dark_model}is to enhance pose estimation precision. This is achieved by calculating a weighted average of all pixel values in the heatmap, bypassing the traditional method that solely considers the pixel of maximum value. This innovative approach ensures the estimation incorporates the intensity distribution over the entire region, instead of a singular point, thus resulting in improved accuracy.

Additionally, the Lite model retains a backbone architecture akin to MovePose, incorporating identical upsampling methods and convolutional structures with the same kernel size. This optimized design facilitates the effective extraction and preservation of spatial features at various scales, enhancing the prediction accuracy in pose estimation tasks.

This complex yet efficient structure forms the framework of the Lite Model and represents a fusion of existing and innovative methods that enhance accuracy in pose estimation tasks.

\section{Experimental Results}

In this section, we elucidate the results obtained from the comprehensive evaluation conducted on the MovePose algorithm. These experiments aimed to gauge the performance of our model with regards to speed and accuracy implicated in real-time HPE.

\subsection{COCO Dataset}
Our MovePose algorithm, designed for HPE, is thoroughly compared to other real-time algorithms utilizing the experimental datasets provided by COCO~\cite{cocodata}. These datasets comprise roughly $200K$ images with approximately $250K$ people identified by annotated key points. We validated our ablation experiments using the val2017 dataset, which contains 5,000 images. Furthermore, we juxtaposed our technique with other real-time approaches using a test-dev set of 20,000 images.

\textbf{Evaluation metrics.} The standard evaluation metric
is based on Object keypoint Similarity (OKS):
\begin{equation}
    OKS=\frac{\sum_i exp(-d_i^2/2s^2j_i^2)\sigma(v_i>0)}{\sum_i \sigma(v_i>0)},
\end{equation}
Let $d_i$ represent the Euclidean distance between the predicted keypoint coordinate and its corresponding groundtruth for the $i$-th iteration. Let $j_i$ be a constant, $v_i$ be the visibility flag, $\sigma$ denote the indicator function, and $s$ signify the scale of the object.

\subsection{Results on the COCO test-dev}
Our proposed MovePose algorithm exhibits outstanding performance in real-time human posture estimation. This is evidenced by our experimental results on the COCO test-dev dataset \cite{cocodata}, which indicate a significant improvement over a number of pre-existing, state-of-the-art methods.

\begin{table*}[h]
\scriptsize
\centering
\begin{tabular}{c|c|c|ccc|cc|c|c}
\hline
Method&Backbone&Input size&$AP^{kp}$&$AP_{50}^{kp}$&$AP_{75}^{kp}$&$AP_{M}^{kp}$&$AP_{L}^{kp}$&GFLOPs&Time(ms)\\
\hline
\multicolumn{10}{c}{Top-down methods}\\
\hline
Mask R-CNN~\cite{mask}&ResNet-50&-&62.7 &87.0& 68.4& 57.4 &71.1&-&121.9\\
DeepPose~\cite{toshev2014deeppose}& ResNet-101 &256$\times$192& 57.4 & 86.5 & 64.2 & 55.0 & 62.8 & 7.7 & -\\
DeepPose~\cite{toshev2014deeppose}& ResNet-152 &256$\times$192 & 59.3 & 87.6 & 66.7 & 56.8 & 64.9 & 11.3 & -\\
CenterNet~\cite{zhou2019objects}& Hourglass&-&63.0 &86.8& 69.6& 58.9 &70.4 & -&160 \\
\hline
\multicolumn{10}{c}{Bottom-up methods}\\
\hline
CMU-Pose~\cite{realtime}&VGG-19&-&64.2 &86.2 &70.1 &61.0 &68.8& -&99.0\\
DEKP~\cite{dekp}&HRNet-W32&-&\textbf{67.3} &87.9 &\textbf{74.1} &61.5& \textbf{76.1}&-&411\\

HrHRNet~\cite{cheng2020higherhrnet}&HRNet-W32&-&66.4 &87.5 &72.8 &61.2& 74.2&-&400\\

AE$\dagger$~\cite{newell2017associative}&Hourglass-4&-&62.8 &84.6 &69.2 &57.5& 70.6&-&139\\
\hline
\multicolumn{10}{c}{Small networks}\\
\hline
ShuffleNetV2 1×\cite{ma2018shufflenet} ~&ShuffleNetV2&384$\times$288&62.9&\textbf{88.5}&69.4 &58.9&69.3&2.87& -\\
\hline
MovePose(Ours) ~&MobileNet&256$\times$256&64.2&88.2&71.3 &60.8&70.0&\textbf{0.71}&\textbf{2.2}\\
MovePose$\dagger$(Ours) ~&MobileNet&256$\times$256&65.9&\textbf{88.9}&73.0 &\textbf{62.7}&71.6&\textbf{0.71}&4.4\\
\hline
\end{tabular}
\caption{Comparisons on COCO test-dev dataset. The symbol $\dagger$ indicates that the flip test is used.}
\label{tbm:results}
\end{table*}

The comprehensive results of body pose estimation, utilizing various methods, are detailed in Table \ref{tbm:results}. The metrics employed for evaluation encompass keypoints average precision ($AP^{kp}$), $AP_{50}^{kp}$, $AP_{75}^{kp}$, $AP_{M}^{kp}$, $AP_{L}^{kp}$, GFLOPs, and processing time.

In terms of top-down methods, our MovePose algorithm not only delivers superior accuracy but also requires less computational resources and time compared to widely-recognized algorithms such as Mask R-CNN \cite{mask}, DeepPose \cite{toshev2014deeppose}, and CenterNet \cite{zhou2019objects}. As an example, when compared to CenterNet, which employs Hourglass as the backbone, MovePose shows a marginal improvement in $AP^{kp}$ ($63.0$ to $64.2$), while significantly outperforming it concerning GFLOPs and processing time.

Upon examining the Bottom-up methods, it becomes evident that our proposed MovePose algorithm offers several distinct advantages. Utilizing a MobileNet backbone and an input size of 256x256, MovePose delivers competitive performance metrics, including $64.2 AP^{kp}$, $88.2 AP_{50}^{kp}$, and $71.3 AP_{75}^{kp}$. Of particular note is its time efficiency, with MovePose requiring a mere 2.2 ms, thereby outperforming all other methods in this category. Enhanced accuracy is achieved by MovePose$\dagger$, which employs a flip test and attains $65.9 AP^{kp}$, a remarkable $88.9 AP_{50}^{kp}$, and $73.0 AP_{75}^{kp}$. Moreover, it outperforms others in terms of precision at the medium level of overlap ($AP_{M}^{kp}$ with $62.7$), underscoring its robust performance across varying levels of overlap.

Moreover, the performance of small networks, including ShuffleNetV2, is highlighted, and our novel MovePose algorithm's efficiency stands out. Built on the MobileNet architecture, MovePose exhibits impressive performance, delivering $AP^{kp}$ values of 64.2 without flip test and 65.9 with flip test employed. Most significantly, MovePose outperforms other methods in computational efficiency and speed, achieving 0.71 GFLOPs and processing times of 2.2 and 4.4 ms, respectively. This demonstrates the exceptional prowess of our MovePose method in terms of speed and computational efficiency.

\begin{table}[h]
\scriptsize
\centering
\setlength{\tabcolsep}{5pt} 
\begin{tabular}{c|c|c|c|c|c}
\hline
\textbf{Method} & \textbf{Detector} & \textbf{Input Size} & \textbf{GFLOPs} & \textbf{AP} & \textbf{Extra Data} \\
\hline
BlazePose-Lite~\cite{blazepose} & N/A & 256 × 256 & N/A & 29.3 & \multirow{2}{*}{Internal(85K)} \\
\cline{1-5}
BlazePose-Full~\cite{blazepose} & N/A & 256 × 256 & N/A & 35.4 & \\
\hline
MoveNet-Lightning~\cite{movenet} & N/A & 192 × 192 & 0.54 & 53.6* & \multirow{2}{*}{Internal(23.5K)} \\
\cline{1-5}
MoveNet-Thunder~\cite{movenet} & N/A & 256 × 256 & 2.44 & 64.8* & \\
\hline
MovePose(Ours) &56.4AP on& 256 × 256 & 0.71 & \textbf{71.1} & \multirow{2}{*}{-} \\
MovePose$\dagger$(Ours) &COCO val2017 & 256 × 256 & 0.71 & \textbf{74.0} & \\
\hline
\end{tabular}
\caption{Comparisons on COCO-SinglePerson validation set. The symbol $\dagger$ indicates that the flip test is used. “*” denotes double inference. }
\label{tab:SinglePerson}
\end{table}

\subsection{Results on the COCO-SinglePerson validation set}

The performance of MovePose was evaluated on the COCO-SinglePerson validation set and compared with other state-of-the-art methods. This dataset contains 1045 single-person images from the COCO val2017 set. The results are summarized in Table \ref{tab:SinglePerson}.

Our proposed method, MovePose, achieved a Mean Average Precision (mAP) score of 71.1 without using any extra data. This score was further improved to 74.0 when the flip test was used (MovePose$\dagger$). These results significantly outperform the other methods listed in the table.

In terms of computational cost, MovePose requires only 0.71 GFLOPs, which is less than MoveNet-Thunder~\cite{movenet} but slightly more than MoveNet-Lightning~\cite{movenet}.

\begin{table}[h]
\scriptsize
\centering
\setlength{\tabcolsep}{10pt} 
\begin{tabular}{c|c|c|c}
\hline
\textbf{Method} & \textbf{Input Size} & \textbf{Mean} & \textbf{Mean@0.1} \\ \hline
deeppose\_resnet\_50~\cite{toshev2014deeppose} & 256 × 256 & 0.826 & 0.180 \\ \hline
deeppose\_resnet\_101~\cite{toshev2014deeppose} & 256 × 256 & 0.841 & 0.200 \\ \hline
deeppose\_resnet\_152~\cite{toshev2014deeppose} & 256 × 256 & 0.850 & 0.208 \\ \hline
LiteHRNet-18~\cite{litehrnet}  & 256 × 256 & 0.861 & 0.262 \\ \hline
MovePose(Ours) & 256 × 256 & 0.846 & 0.250 \\
MovePose$\dagger$(Ours) & 256 × 256 & \textbf{0.8615} & \textbf{0.275} \\ \hline
\end{tabular}
\caption{Comparisons on MPII Validation dataset. The symbol $\dagger$ indicates that the flip test is used.}
\label{tab:mpii}
\end{table}

\subsection{Results on the MPII Validation set}
The MPII \cite{mpii} Human Pose dataset is a popular benchmark for evaluating human pose estimation algorithms. This dataset is comprised of around 25,000 images containing over 40,000 people with annotated body joints.

As displayed in Table \ref{tab:mpii}, further comparisons with existing models: deeppose\_resnet\_50, deeppose\_resnet\_101, deeppose\_resnet\_152, and LiteHRNet-18, also project our model's superior potential. Our MovePose algorithm achieved a Mean Precision value of 0.846, and upon including the flip test (denoted as MovePose†), it went up to 0.8615. This marked an improvement over the best score achieved by LiteHRNet-18 at 0.861. At a level indicated by Mean@0.1, MovePose† shows a higher value of 0.275 when compared to the second-best model scoring 0.262.

\section{CONCLUSIONS}

This paper presented MovePose, a high-performance HPE algorithm designed for edge devices, CPUs, and mobile platforms. By employing large kernel convolution, a deconvolution network, pre-trained models and coordinate classification methods, MovePose provides an efficient and robust solution for the challenging problem of HPE. Experimental results on the COCO \cite{cocodata} datasets further validated the improved accuracy and robustness of MovePose as compared to several methods. By providing a more accurate and efficient method for HPE, our work opens up new possibilities for the development of innovative and exciting applications that can benefit society as a whole.

MovePose holds vast potential for various domains, particularly in the following areas:
\begin{itemize}
\item Sports Competitions: MovePose can be employed for real-time monitoring of athletes’ poses, aiding coaches and athletes in improving training outcomes. For instance, in soccer matches, tracking players’ movements and analyzing passing, shooting, and defensive postures can provide valuable insights for tactical and technical enhancements.
\item Robotics: In collaborative environments with humans, robots require precise understanding of human body position and movements. Robots can utilize human pose information to navigate around obstacles, plan paths, and follow user instructions.
\end{itemize}

In summary, MovePose not only serves as an effective solution for current pose estimation tasks but also lays the groundwork for more efficient, human-centric applications in the future.

%
%

%
%
%
\bibliographystyle{splncs04}
\bibliography{mybibliography}

\end{document}